# A Class of DSm Conditioning Rules[1]

Florentin Smarandache, Mark Alford

Air Force Research Laboratory, RIEA,

525 Brooks Rd., Rome, NY 13441-4505, USA

**Abstract**:

In this paper we introduce two new DSm fusion conditioning rules with example, and as a generalization of them a class of DSm fusion conditioning rules, and then extend them to a class of DSm conditioning rules.

## 0. Introduction

In order to understand the material in this paper, it is first necessary to define the terms that we'll be using:

- Frame of discernment = the set of all hypotheses.
- Ignorance is the mass (belief) assigned to a union of hypotheses.
- Conflicting mass is the mass resulted from the combination of many sources of information of the hypotheses whose intersection is empty.
- Fusion space = is the space obtained by combining these hypotheses using union, intersection, or complement – depending on each fusion theory.
- Dempster-Shafer Theory is a fusion theory, i.e. method of examination of hypotheses based on measures and combinations of beliefs and plausibility in each hypothesis, beliefs provided by many sources of information such as sensors, humans, etc.
- Transferable Belief Model is also a fusion theory, an alternative of DST, whose method is of transferring the conflicting mass to the empty set.

---

[1] This research has been supported by Air Force Research Laboratory, Rome, NY, USA, in June and July 2009.

- Dezert-Smarandache Theory is a fusion theory, which is a natural extension of DST and works for high conflicting sources of information, and overcomes the cases where DST doesn't work.
- Power set = is the fusion space of Dempster-Shafer Theory (DST) and Transferable Belief Model (TBM) theory; the power set is the set of all subsets of the frame of discernment, i.e. all hypotheses and all possible unions of hypotheses. {In the fusion theory union of hypotheses means uncertainty about these hypotheses.}
- Hyper-power set = the fusion set of Dezert-Smarandache Theory (DSmT); the hyper-power set is the set formed by all unions and intersections of hypotheses. {By intersection of two or more hypotheses we understand the common part of these hypotheses – if any. In the case when their intersection is empty, we consider these hypotheses disjoint.}
- Super-power set = the fusion space for the Unification of Fusion Theories and rules; the super-power set is the set formed by all unions, intersection, and complements of the hypotheses. {By a complement of a hypothesis we understand the opposite of that hypothesis.}
- Basic belief assignment (bba), also called mass and noted by m(.), is a subjective probability (belief) function that a source assigns to some hypotheses or their combinations. This function is defined on the fusion space and whose values are in the interval [0, 1].

In the first section, we consider a frame of discernment and then we present the three known fusion spaces. The first fusion space, the power set, is used by Dempster-Shafer Theory (DST) and the Transferable Believe Model (TBM). The second fusion space, which is larger, the hyper-power set, is used by Dezert-Smarandache Theory (DSmT), while the third fusion space, the super-power set, is the most general one, and it is used in the Unification of Fusion Theories and Rules.

In the second section we present Dempster's conditioning rule and the Bel(.) and Pl(.) functions.

In order to overcome some difficult corner cases where Dempster's Conditioning Rule doesn't work, we design the first simple DSm conditioning rule and the second simple DSm conditioning rule in section 3. These rules are referring to the fact that: if a source provides us some evidence (i.e. a basic belief assignment), but later we find out that the true hypothesis is in a subset A of the fusion space, then we need to compute the conditional belief m(.|A).

In section 4 we give a Class of DSm Conditioning Rules that generalizes two simple DSm conditioning rules cover.

In section 5 we present two examples in military about target attribute identification.

2 This research has been supported by Air Force Research Laboratory, Rome, NY, USA, in June and July 2009.

## 1. Mathematical Preliminaries.

Let $\Theta = \{\theta_1, \theta_2, \ldots, \theta_n\}$, with $n \geq 2$, be a frame of discernment.

As fusion space, Shafer uses the power set $2^\Theta$, which means $\Theta$ closed under union of sets, $(\Theta, \cup)$, and it is a lattice. In Dempster-Shafer Theory (DST) all hypotheses $\theta_i$ are considered mutually exclusive, i.e. $\theta_i \cap \theta_j = \phi$ for any $i \neq j$, and exhaustive.

Dezert extended the power set to a hyper-power set $D^\Theta$ in Dezert-Smarandache Theory (DSmT), which means $\Theta$ closed under union and intersection of sets $(\Theta, \cup, \cap)$ and it is a distribute lattice; in this case the hypotheses are not necessarily exclusive, so there could be two or more hypotheses whose intersections are non-empty. Each model in DSmT is characterized by empty and non-empty intersections. If all intersections are empty, we get Shafer's model used in DST; if some intersection are empty and others are not, we have a hybrid model; and if all intersection are non-empty we have a free model.

Further on Smarandache [3] extended the hyper-power to a super-power set $S^\Theta$, as in UFT (Unification of Fusion Theories), which means $\Theta$ closed under union, intersection, and complement of sets $(\Theta, \cup, \cap, \mathcal{C})$, that is a Boolean algebra.

We note by G any of these three fusion spaces, power set, hyper-power set, or super-power set.

## 2. Dempster's Conditioning Rule (DCR).

Let's have a bba (basic believe assignment, also called mass):

$$m_1: G^\Theta \rightarrow [0, 1], \text{ where } \sum_{X \in G^\Theta} m_1(X) = 1.$$

In the main time we find out that the truth is in $B \in G^\Theta$. We therefore need to adjust our bba according to the new evidence, so we need to compute the conditional bba $m_1(X|B)$ for all $X \in G^\Theta$.

Dempster's conditioning rule means to simply fuse the mass $m_1(.)$ with $m_2(B) = 1$ using Dempster's classical fusion rule.

A similar procedure can be done in DSmT, TBM, etc. by combining $m_1(.)$ with $m_2(B) = 1$ using other fusion rule.

In his book Shafer gave the conditional formulas for believe and plausible functions Bel(.) and respectively Pl(.) only, not for the mass m(.).

In general we know that:

3 | This research has been supported by Air Force Research Laboratory, Rome, NY, USA, in June and July 2009.

$$\text{Bel}(A) = \sum_{X \subseteq A} m_1(X)$$

and

$$\text{Pl}(A) = \sum_{X \cap A \neq \phi} m_1(X).$$

Let $m_1(.)$ and $m_2(.)$ be two bba's defined on $G^\Theta$. The conjunctive rule for combining these bba's is the following:

$$(m_1 + m_2)(A) = \sum_{\substack{X, Y \in G^\Theta \\ X \cap Y = A}} m_1(X) m_2(Y)$$

In order to compute in DST the subjective conditional probability of B given A, i.e. $m(A|B)$, Shafer combines the masses $m_1(.)$ and $m_2(B)=1$ using Dempster's rule (pp. 71-72 in [2]) and he gets:

$$m(A|B) = \frac{\sum_{X \cap Y = A} m_1(X) m_2(Y)}{1 - \sum_{X \cap Y = \phi} m_1(X) m_2(Y)} \quad \text{(which is Dempster's rule)}$$

$$= \frac{\sum_{X \cap B = A} m_1(X) m_2(B)}{1 - \sum_{X \cap B = \phi} m_1(X) m_2(B)} \quad \text{(since only } m_2(B) \neq 0, \text{ all other values of } m_2(Y) = 0 \text{ for } Y \neq B)$$

$$= \frac{\sum_{X \cap B = A} m_1(X)}{1 - \sum_{X \cap B = \phi} m_1(X)} = \frac{\sum_{X \cap B = A} m_1(X)}{\sum_{X \cap B \neq \phi} m_1(X)} \quad \text{which is exactly what Milan Daniel got in [1], but with different notations.}$$

Therefore, **Dempster's Conditioning Rule** (DCR) referred to masses {not to $\text{Bel}(.)$ or to $\text{Pl}(.)$ functions as designed by Shafer} is the following:

$$\forall A \in 2^\Theta \setminus \phi \text{ we have } m_{\text{DCR}}(A|B) = \frac{\sum_{X \cap B = A} m_1(X)}{\sum_{X \cap B \neq \phi} m_1(X)}.$$

With M. Daniel's notations, Dempster's Conditioning Rule becomes:

4 | This research has been supported by Air Force Research Laboratory, Rome, NY, USA, in June and July 2009.

$\forall\, X \in D^{\Theta} \setminus \phi$ we have $m_{DCR}(X|A) = \dfrac{\sum\limits_{Y \cap A = X} m_1(Y)}{\sum\limits_{Y \cap A \neq \phi} m_1(Y)}$.

DCR doesn't work when Pl(A) = 0 since its denominator becomes null.

### 3. Two DSm Conditioning Rules.

We can overcome this undefined division by constructing a **DSm first simple conditioning rule** in the super-power set:

$\forall\, X \in S^{\Theta} \setminus \phi$ we have $m_{DSmT1}(X|A) = \sum\limits_{(Y \cap A = X)\, or\, (Y \cap A = \phi\, and\, X = A)} m(Y)$

which works in any case.

In the corner case when Pl(A) = 0, we get $m_{DSmT1}(A|A) = 1$ and all other $m_{DSmT1}(X|A) = 0$ for $X \neq A$.

The DSm first simple conditioning rule transfers the masses which are outside of A (i.e. the masses m(Y) with $Y \cap A = \phi$) to A in order to keep the normalization of m(.), in order to avoid doing normalization by division as DCR does.

Another way will be to uniformly split the total mass which is outside of A:

$$K_{cond} = \sum\limits_{Y \cap A = \phi} m(Y)$$

to the non-empty sets of $\mathcal{P}(A)$, i.e. sets whose mass is non-zero, where $\mathcal{P}(A)$ is the set of all parts of A.

So, a **DSm second simple conditioning rule** is:

$m_{DSmT2}(X\,|\,A) = \sum\limits_{Y \cap A = X} m(Y) + \dfrac{1}{C_{P(A)}} \cdot \sum\limits_{Y \cap A = \phi} m(Y)$

where $C_{P(A)}$ is the cardinal of the set of elements from P(A) whose masses are not zero, i.e.

$C_{P(A)} = \mathrm{Card}\{Z\,|\,Z \in S^{\Theta},\, Z \subseteq A,\, \sum\limits_{Y \cap A = Z} m(Y) \neq 0\}$.

---

**5** This research has been supported by Air Force Research Laboratory, Rome, NY, USA, in June and July 2009.

In the corner edge when $C_{P(A)} = 0$, we replace it with the number of singletons included in A if any, the number of unions of singletons included in A if any, and A itself.

## 4. A Class of DSm Conditioning Rules.

In this way we can design a **class of DSm conditioning rules** taking into consideration not only masses, but also other parameters that might influence the decision-maker in calculating the subjective conditioning probability, and which is a generalization of Dempster's conditioning rule:

$$m_{DSmTclass}(X \mid A) = \frac{\sum_{Y \cap A = X} \frac{\alpha(Y)}{\beta(Y)}}{\sum_{Y \cap A = \phi} \frac{\alpha(Y)}{\beta(Y)}}$$

with $\alpha(Y) = \alpha_1(Y) \cdot \alpha_2(Y) \cdot \ldots \cdot \alpha_p(Y)$, where all $\alpha_i(Y)$, $1 \leq i \leq p$, are parameters that Y is directly proportional to;

and $\beta(Y) = \beta_1(Y) \cdot \beta_2(Y) \cdot \ldots \cdot \beta_r(Y)$, where all $\beta_j(Y)$, $1 \leq j \leq r$, are parameters that Y is inversely proportional to.

## 5. Examples of Conditioning Rules.

**Example 5.1**.

Let $m_1(.)$ be defined on the frame {F = friend, E = enemy, N = neutral}, where the hypotheses F, E, N are mutually exclusive, in the following way (see the second row):

|  | $\phi$ | F | E | N | F∪E | F∪E∪N | N∩(F∪E) |
|---|---|---|---|---|---|---|---|
| $m_1$ | 0 | 0.2 | 0.1 | 0.3 | 0.1 | 0.3 | 0 |
| $m_2$ | 0 | 0 | 0 | 0 | 1 | 0 | 0 |
| $m_1 + m_2$ | 0 | 0.2 | 0.1 | 0 | 0.4 | 0 | 0.3 |
| $m_{DCR}(X\mid F\cup E)$ | 0 | 2/7 | 1/7 | 0 | 4/7 | 0 | 0 |
| $m_{TBM}(X\mid F\cup E)$ | 0.3 | 0.2 | 0.1 | 0 | 0.4 | 0 | 0 |
| $m_{DSmT1}(X\mid F\cup E)$ | 0 | 0.2 | 0.1 | 0 | 0.7 | 0 | 0 |
| $m_{DSmT2}(X\mid F\cup E)$ | 0 | 0.3 | 0.2 | 0 | 0.5 | 0 | 0 |

Table 1

6 This research has been supported by Air Force Research Laboratory, Rome, NY, USA, in June and July 2009.

Suppose the truth is in the set F∪E. First we combine $m_1(.)$ with $m_2(E) = 1$ using the conjunctive rule, and its result $m_1 + m_2$ is in the fourth row in Table 1. All below conditioning rules are referred to the result of this conjunctive rule, and they differ through the way the conflicting mass, i.e. mass of empty intersections, is transferred to the other elements.

In DCR, since N∩ (F∪E) = φ the conflicting mass $m_1(N) \cdot m_2(F \cup E) = 0.3 \cdot 1 = 0.3$, is transferred to the non-empty sets F, E, and F∪E proportionally with respect to their masses acquired after applying the conjunctive rule ($m_1 + m_2$), i.e. with respect to 0.2, 0.1, and respectively 0.4. Thus, we get $m_{DST}(X|F \cup E)$ as in the fifth row of Table 1, where X ∈ { φ, F, E, N, F∪E, F∪E∪N, N∩ (F∪E)}.

In Smets' TBM (Transferable Believe Model), the conflicting mass, 0.3, is transferred to the empty set, since TBM considers an open world (non-exhaustive hypotheses). See row # 6.

With DSm first conditioning rule (row # 7) the conflicting mass 0.3 is transferred to the whole set that the truth belongs to, F∪E. So, $m_{DSmT1}(F \cup E | F \cup E) = (m_1 + m_2)(F \cup E) + 0.3 = 0.4 + 0.3 = 0.7$.

In DSm second conditioning rule (row # 8) the conflicting mass 0.3 is uniformly transferred to the non-empty sets F, E, and F∪E, therefore each such set receives $0.3/3 = 0.1$.

**Example 5.2.**

Let $m_1(.)$ be defined on the frame {A = Airplane, T = tank, S = ship, M = submarine}, where the hypotheses A, T, S, M are mutually exclusive, in the following way (see the second row):

| | φ | A | T | S | M | A∪S | T∪M | A∩(T∪M) | S∩(T∪M) | (A∪S)∩(T∪M) |
|---|---|---|---|---|---|---|---|---|---|---|
| $m_1$ | 0 | 0.4 | 0 | 0.5 | 0 | 0.1 | 0 | | | |
| $m_2$ | 0 | 0 | 0 | 0 | 0 | 0 | 1 | | | |
| $m_1 + m_2$ | 0 | 0 | 0 | 0 | 0 | 0 | 0 | 0.4 | 0.5 | 0.1 |
| $m_{DCR}(X|T \cup M)$ | 0 | N/A | N/A | N/A | N/A | N/A | N/A | | | |
| $m_{TBM}(X|T \cup M)$ | 1 | 0 | 0 | 0 | 0 | 0 | 0 | 0 | 0 | 0 |
| $m_{DSmT1}(X|T \cup M)$ | 0 | 0 | 0 | 0 | 0 | 0 | 1 | 0 | 0 | 0 |
| $m_{DSmT2}(X|T \cup M)$ | 0 | 0 | 1/3 | 0 | 1/3 | 0 | 1/3 | 0 | 0 | 0 |

Table 2

Suppose the truth is in T∪M. Since the sets A∩ (T∪M), S∩ (T∪M), and (A∪S)∩ (T∪M) are empty, their masses 0.4, 0.5, and respectively 0.1 have to be transferred to non-empty sets belonging to P(T∪M), where P(T∪M) means the set of all subsets of T∪M.

7 This research has been supported by Air Force Research Laboratory, Rome, NY, USA, in June and July 2009.

In this case, DCR does not work since it gets an undefined division 0/0.

In Smets' TBM (Transferable Believe Model), the total conflicting mass, $0.4 + 0.5 + 0.1 = 1$, is transferred to the empty set, since TBM considers an open world (non-exhaustive hypotheses). See row # 6.

With DSm first conditioning rule (row # 7) the total conflicting mass, 1, is transferred to the whole set that the truth belongs to, $T \cup D$. So, $m_{DSmT1}(T \cup D | T \cup D) = (m_1 + m_2)(T \cup D) + 1 = 1$.

In DSm second conditioning rule (row # 8) the total conflicting mass is 1. Since $C_{(B \cup D)} = 0$, the total conflicting mass 1 is uniformly transferred to the sets T, D, and $T \cup D$ {i.e. the singletons and unions of singletons included in $T \cup D$}, therefore each such set receives 1/3.

**Conclusion**.

We have examined Dempster's Conditioning Rule in terms of bba. We saw that in the second military example, using DCR for target identification, the procedure failed mathematically. That's why we designed two DSm simple conditioning rules and could complete the procedure of target identification. We have compared these approaching of target identification using DCR, TBM conditioning, and the two DSm conditioning rules that got better results than DCR and TBM. We also observed from these examples that the two DSm simple conditioning rules give almost similar results.

8 | This research has been supported by Air Force Research Laboratory, Rome, NY, USA, in June and July 2009.

9 This research has been supported by Air Force Research Laboratory, Rome, NY, USA, in June and July 2009.